\pdfoutput=1
\documentclass{article}
\usepackage{aaai25}  
\usepackage{times}  
\usepackage{helvet}  
\usepackage{courier}  
\usepackage[hyphens]{url}  
\usepackage{graphicx} 
\usepackage{amsmath}
\usepackage{amssymb}
\usepackage[square,sort,comma,numbers]{natbib}
\usepackage{algorithm}
\usepackage{algpseudocode}
\usepackage{caption}
\usepackage{tabularx}
\usepackage{booktabs}
\usepackage{multirow}
\usepackage{array}
\usepackage{xcolor}
\usepackage{svg}
\usepackage{newfloat}
\usepackage{listings}

\usepackage{tabularx}
\usepackage{booktabs}
\usepackage{multirow}
\usepackage{array}
\usepackage{xcolor}  
\usepackage{svg}

\DeclareCaptionStyle{ruled}{labelfont=normalfont,labelsep=colon,strut=off} 
\floatstyle{ruled}
\newfloat{listing}{tb}{lst}{}
\floatname{listing}{Listing}

\title{DH-RAG: A Dynamic Historical Context-Powered Retrieval-Augmented Generation Method for Multi-Turn Dialogue}
\author{
    Feiyuan Zhang\textsuperscript{1},
    Dezhi Zhu\textsuperscript{2},
    James Ming\textsuperscript{3},
    Yilun Jin\textsuperscript{1},
    Di Chai\textsuperscript{1},
    Liu Yang\textsuperscript{1},
    Han Tian\textsuperscript{4},
    Zhaoxin Fan\textsuperscript{5,*},
    Kai Chen\textsuperscript{1,*}
}
\affiliations{
    \textsuperscript{1}Hong Kong University of Science and Technology\\
    \textsuperscript{2}Huazhong University of Science and Technology\\
    \textsuperscript{3}University of California San Diego\\
    \textsuperscript{4}University of Science and Technology of China\\
    \textsuperscript{5}Beihang University
}

\nocopyright  

\begin{document}

\maketitle

\begin{abstract}
Retrieval-Augmented Generation (RAG) systems have shown substantial benefits in applications such as question answering and multi-turn dialogue \citep{lewis2020retrieval}. However, traditional RAG methods, while leveraging static knowledge bases, often overlook the potential of dynamic historical information in ongoing conversations. To bridge this gap, we introduce DH-RAG, a Dynamic Historical Context-Powered Retrieval-Augmented Generation Method for Multi-Turn Dialogue. DH-RAG is inspired by human cognitive processes that utilize both long-term memory and immediate historical context in conversational responses \citep{stafford1987conversational}. DH-RAG is structured around two principal components: a History-Learning based Query Reconstruction Module, designed to generate effective queries by synthesizing current and prior interactions, and a Dynamic History Information Updating Module, which continually refreshes historical context throughout the dialogue. The center of DH-RAG is a Dynamic Historical Information database, which is further refined by three strategies within the Query Reconstruction Module: Historical Query Clustering, Hierarchical Matching, and Chain of Thought Tracking. Experimental evaluations show that DH-RAG significantly surpasses conventional models on several benchmarks, enhancing response relevance, coherence, and dialogue quality.
\end{abstract}

\vspace{-0.2in}

\section{Introduction}

Dialogue systems and question-answering tasks have increasingly captured the interest of the artificial intelligence community. With the evolution of Large Language Models (LLMs) \cite{achiam2023gpt,touvron2023llama}, Retrieval-Augmented Generation (RAG) methods have showcased significant benefits in these areas. RAG systems, by amalgamating external knowledge bases with generative models, furnish responses not only more precise but also richly informative. As an indispensable adjunct to LLMs, RAG systems are instrumental in elevating the conversational quality of these robust models.

\begin{figure}[t]
    \centering
    \includegraphics[width=0.45\textwidth]{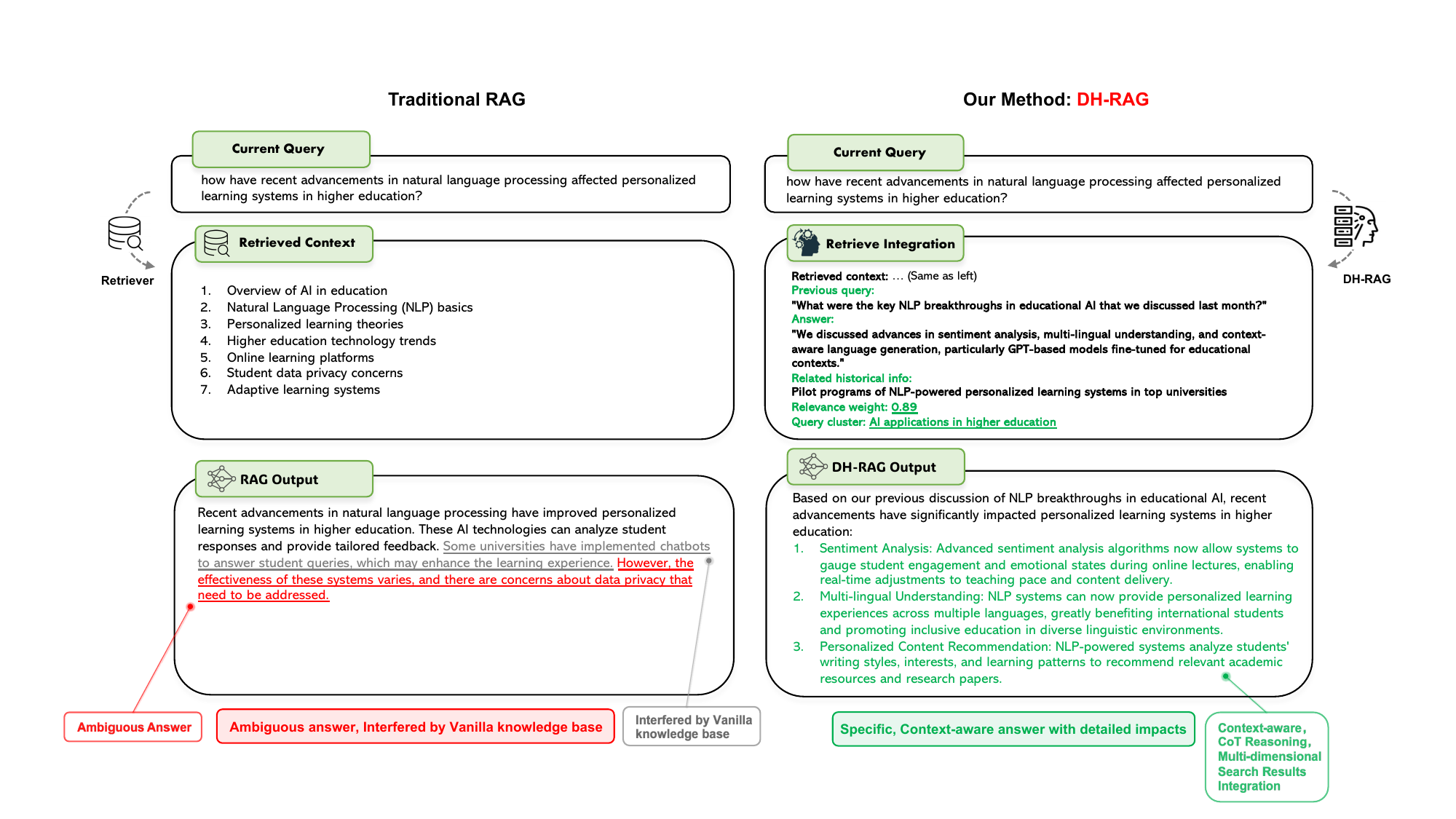}
    \caption{Difference between existing RAG methods and our DH-RAG method.}
    \label{fig:dh-rag-comparison}
\end{figure}

\vfill
\begin{flushleft}
\footnotesize
\textit{This paper is accepted by AAAI'25 Workshop on Agents for Information Retrieval (Agent4IR).\\
The extended version is available at arXiv: https://arxiv.org/abs/2403.xxxxx}
\end{flushleft}

\newpage

The pioneering work by \cite{lewis2020retrieval} introduced Retrieval-Augmented Generation (RAG) models for LLMs, which ingeniously combine pre-trained parametric and non-parametric memory for language generation. Subsequent research \cite{gao2023retrieval, jiang2023active, chen2024benchmarking, yu2022retrieval} proposed enhancements to RAG models, focusing on the aspects of retrieval \cite{wang2023knowledgpt, cheng2024lift} and generation \cite{anderson2022lingua, jiang2023longllmlingua}. Despite achieving remarkable performance, existing RAG methods predominantly utilize static knowledge bases, thus neglecting the critical role of dynamically updated and utilized historical information in multi-turn dialogues. However, it is a well-acknowledged fact that human cognitive processes harness both long-term memory and immediate historical context in generating conversational responses \cite{xu2021beyond,stafford1987conversational}.

Specifically, in human interactions, responses are influenced not only by long-term memory, comparable to static knowledge bases, but also by short-term dynamic historical information crucial for contextual understanding and appropriate response formulation. This dynamic component is vital for maintaining dialogue coherence and relevance. However, existing RAG systems often fail to effectively harness these dynamically evolving contextual cues in multi-turn dialogues, possibly leading to responses that are not well-integrated with the overall conversational flow. This limitation motivates a pivotal question: \textbf{\emph{Can we develop a RAG model that effectively utilizes both static external knowledge and the transient context inherent in ongoing conversations?}}

To address this issue, we introduce DH-RAG, a \textbf{D}ynamic \textbf{H}istorical Context-Powered \textbf{R}etrieval-\textbf{A}ugmented \textbf{G}eneration method tailored for multi-turn dialogue, as illustrated in Fig. \ref{fig:dh-rag-comparison}. Developed under the principle that dynamic context is crucial for query construction, DH-RAG mimics the human cognitive process of integrating long-term memory with short-term dynamic information during conversations. To realize this, we propose two innovative modules: the History-Learning based Query Reconstruction Module and the Dynamic History Information Updating Module. The former is designed to generate more effective queries by synthesizing both current and historical queries, while the latter updates historical information in real-time throughout the conversation, employing a time-based and similarity-based elimination strategy. Furthermore, the success of DH-RAG heavily relies on a Dynamic Historical Information database within the History-Learning based Query Reconstruction Module. To optimize its effectiveness, we introduce three strategic approaches: Historical Query Clustering, Hierarchical Matching, and Chain of Thought Tracking, which collectively enhance the construction of the database. These designs collectively endow DH-RAG with a robust capability to enhance conversation quality.

To ascertain the efficacy of DH-RAG, we construct a benchmark leveraging several widely recognized datasets. Comprehensive experiments illustrate that DH-RAG consistently surpasses conventional models and secures state-of-the-art performance.

Our main contributions  can be summarized as:

\begin{itemize}
    \item We introduce the DH-RAG method, marking the first endeavor to integrate a dynamic historical information processing mechanism within the RAG framework.
    \item We develop a History-Learning based Query Reconstruction Module and a Dynamic History Information Updating Module, significantly enhancing the effectiveness of information utilization in multi-turn dialogues.
    \item We implement three innovative strategies: Historical Query Clustering, Hierarchical Matching, and Chain of Thought Tracking, all aimed at augmenting the system's performance.
    \item We perform rigorous experiments on multiple benchmark datasets, which show that DH-RAG markedly surpasses existing methods in terms of response relevance, coherence, and overall dialogue quality.
\end{itemize}

%

\section{Related Work}
\subsection{Retrieval-Augmented Generation (RAG)}
Retrieval-Augmented Generation (RAG) systems significantly advance the capabilities of dialogue systems and question-answering tasks by amalgamating external knowledge bases with generative models. \cite{lewis2020retrieval} introduces the RAG models, adeptly merging pre-trained parametric and non-parametric memories for enhanced language generation. Subsequent studies \cite{liu2020retrieval} introduce several enhancements to RAG models, focusing on refining retrieval \cite{wang2023knowledgpt, cheng2024lift} and enhancing generation capabilities \cite{anderson2022lingua, jiang2023longllmlingua}. Recent innovations include FLARE \cite{zhang2023flare}, which introduces a feedback loop augmented retrieval method to iteratively refine retrieval outcomes and bolster generation quality. Additionally, SelfRAG \cite{asai2023selfrag} presents a self-supervised retrieval-augmented framework that boosts both retrieval and generation processes through the strategic use of pseudo-labels generated by the model itself. Despite these significant advancements, the challenge of seamlessly integrating dynamic historical context in RAG models for multi-turn dialogues remains an elusive goal.

Though achieve remarkable progress, most existing approaches continue to depend predominantly on static knowledge bases and do not adequately address the need to capture the evolving contextual nuances within conversations. This gap propels the development of DH-RAG in this paper, aimed at more effectively incorporating both static external knowledge and the transient context prevalent in ongoing dialogues, thereby enhancing the quality and coherence of multi-turn dialogue interactions.

\subsection{Retrieval-based Dialogue Systems} 

Retrieval-based dialogue systems \cite{ni2023recent,tao2021survey} become central in natural language processing, aiming to generate responses from large conversational datasets \cite{achiam2023gpt}. Early models like the reformulation-based retrieval system by \cite{yan2016learning} improve response matching in information-seeking dialogues. Subsequent developments include multi-hop retrieval methods \cite{xu2021beyond} that navigate complex knowledge graphs for richer responses and few-shot learning approaches like \cite{yu2021few}, which enhance performance in resource-scarce scenarios. However, these systems often struggle with generating context-appropriate responses in multi-turn dialogues due to their reliance on selecting pre-existing responses. This has led to the exploration of hybrid models \cite{sawarkar2024blended,xu2024retrieval}, such as the selective knowledge fusion framework by \cite{su2023selective}, which increases response relevance and coherence.

Despite progress, smoothly integrating changing conversational contexts remains challenging. This paper introduces DH-RAG, a novel framework that addresses this gap in multi-turn dialogue systems by creating a dynamic historical information database and a History-Learning Based Query Reconstruction Module. This approach refines queries using past context, enhancing dialogue interaction quality.

\subsection{Long-term and Short-term Memory Theory} 

The study of long-term and short-term memory is essential in cognitive psychology, providing key insights into how humans process information and make decisions, and have inspired many researcher in RAG \cite{cheng2024lift,hu2023reveal} and dialogue system fields \cite{zhang2019memory, bang2015example}. The foundational work by \cite{atkinson1968human} introduced the multi-store model of memory, differentiating between short-term and long-term memory systems. This model has been refined over the years, notably by \cite{baddeley1974working}, who developed the working memory model that focuses on the active management of information in short-term storage. Recent studies have examined the interaction between these memory systems. For example, \cite{dudai2015consolidation} investigates the processes of memory consolidation and reconsolidation. In conversational contexts, \cite{horton2016conversational} showed how both long-term and short-term memory help in forming a shared understanding in dialogues. The work of \cite{kumar2022role} further highlights the importance of working memory in keeping conversations coherent over multiple turns.

In this paper, we are inspired by a common sense notion that both long-term memory from a static database and short-term memory from current dialogues should serve as important contexts in the RAG process. Therefore, we propose the DH-RAG method, which skillfully utilizes the dynamic history context (short-term memory) through two novel modules and a newly constructed database to enhance the RAG process.

\section{Method}

\begin{figure*}
    \centering
    \includegraphics[width=1.05\textwidth]{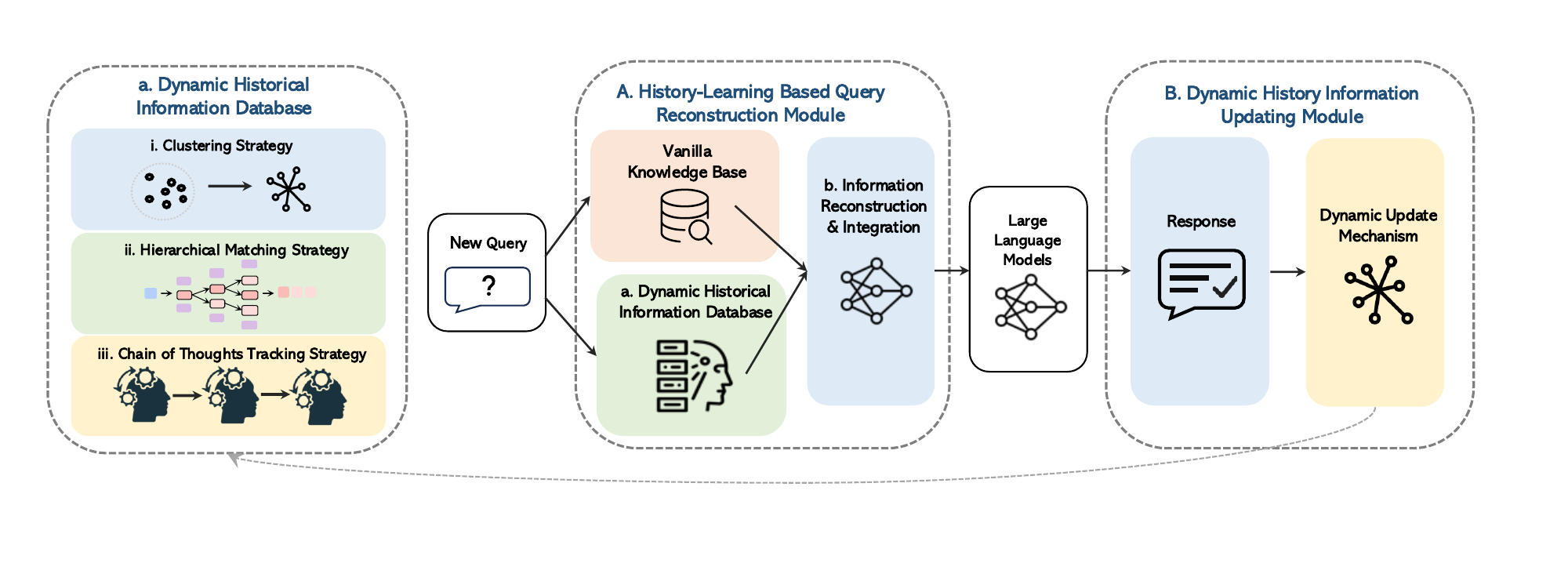}
       \vspace{-0.5in}
    \caption{Overall pipeline of our DH-RAG method.}
    \label{fig:dh-rag-framework}
    \vspace{-0.2in}
\end{figure*}

\subsection{Overview}

\begin{algorithm}
\caption{DH-RAG Framework}
\label{alg:DH-RAG}
\begin{algorithmic}[1]
\Require Query $q_{\text{new}}$, Knowledge Base $\mathcal{D}$, Historical Database $\mathcal{H}$
\Ensure Response $r$
\footnotesize 
\State $R_v \gets \text{VanillaRetrieval}(q_{\text{new}}, \mathcal{D})$ \Comment{Vanilla Retrieve}
\State \textbf{// Clustering Matching}
\State $\mathcal{C} \gets \text{ClusterHistoricalQueries}(\mathcal{H})$
\State \textbf{// Hierarchical Matching}
\State $S_c \gets \text{CategoryMatching}(q_{\text{new}}, \mathcal{C})$ \Comment{Dynamic Retrieve}
\State $R_h \gets \emptyset$
\For{$c_i \in S_c$}
    \State $S_h \gets \text{QueryMatching}(q_{\text{new}}, c_i)$
    \For{$h_j \in S_h$}
        \State $R_h \gets R_h \cup \text{ExtractInfo}(h_j)$
    \EndFor
\EndFor
\State \textbf{// Chain of Thoughts Matching}
\State $S_g \gets \text{TopKSimilarQueries}(q_{\text{new}}, \mathcal{H})$
\State $R_g \gets \emptyset$
\For{$h_i \in S_g$}
    \State $R_g \gets R_g \cup \text{ExtractCoT}(h_i)$
\EndFor
\State $R_{\text{final}} \gets \text{InformationIntegration}(R_v, R_h, R_g)$
\State $r \gets \text{Generate}(q_{\text{new}}, R_{\text{final}})$ \Comment{Generate}
\State \textbf{// Dynamic Update}
\State $\mathcal{H}' \gets \mathcal{H} \cup \{(q_{\text{new}}, r)\}$ \Comment{Update}
\State $\mathcal{H}_{\text{updated}} \gets \text{FilterAndScore}(\mathcal{H}')$
\State $\mathcal{C} \gets \text{UpdateClusters}(\mathcal{H}_{\text{updated}})$
\State \Return $r$
\end{algorithmic}
\end{algorithm}

In this study, we explore multi-turn dialogue systems and retrieval augmented generation. Traditional Retrieval-Augmented Generation (RAG) systems often struggle with the continuity and context dependency inherent in multi-turn dialogues, where user queries evolve or deviate based on preceding interactions. To address this challenge, we introduce the Dynamic History-aware Retrieval-Augmented Generation (DH-RAG) system. This innovative approach is specifically tailored to manage and integrate sequences of contextually interconnected queries effectively.

Before we explain our system, let's define the problem. Consider a series of queries $Q = \{q_1, q_2, ..., q_t\}$, where $q_t$ is the current query. Our goal is to generate a response $r_t$. We can describe this process with the equation:

\begin{equation}
    r_t = f(q_t, K, H)
\end{equation}

Here, $K$ is the static knowledge base that provides background information, and $H = \{(q_1, p_1, r_1), (q_2, p_2, r_2), ..., (q_{t-1}, p_{t-1}, r_{t-1})\}$ represents the history of past queries, passages, and responses. $p_i$ is the passage retrieved for query $q_i$. The challenge is to create a function $f$ that uses the current query $q_t$, the knowledge $K$, and the history $H$ to generate a suitable response $r_t$.

Figure \ref{fig:dh-rag-framework} illustrates the workflow of our system during a multi-turn conversation. The process begins with the system receiving a new query from the user. This query is first processed by the History-Learning Based Query Reconstruction Module, which leverages both a standard knowledge base and a Dynamic Historical Information Database to enrich and integrate pertinent information. The enhanced query information is subsequently passed to a LLM, which generates the user's response. Finally, the Dynamic History Information Updating Module updates the Dynamic Historical Information Database with the new response, ensuring that the system evolves and remains relevant with each interaction.  Algorithm \ref{alg:DH-RAG} show the overall pipeline of  DH-RAG. Our method comprises two key modules and a dedicated database: the History-Learning Based Query Reconstruction Module,  the
 Dynamic History Information Updating Module, and the {Dynamic Historical Information Database. Next, we introduce the two modules and the Dynamic Historical Information Database in detail.

    
    

Next, we introduce the two modules and the Dynamic Historical Information Database in detail.

\subsection{History-Learning based Query Reconstruction Module}

Given a query, it is crucial to merge static knowledge with dynamic historical information to craft a new query that informs the subsequent response generated by a LLM. This approach introduces a novel concept in the retrieval process. To facilitate this, we introduce the \emph{History-Learning Based Query Reconstruction Module}, which is designed to fulfill this specific objective. This module comprises three key components: the \emph{Vanilla Static Knowledge Base}, the \emph{Dynamic Historical Information Database}, and the \emph{Information Reconstruction \& Integration Process}. 

\subsubsection{Vanilla Static Knowledge Base}

Traditional RAG systems fundamentally depend on a standard knowledge base. In our DH-RAG, we continue to utilize a static knowledge base as the foundational element. When presented with a query $q$, we retrieve a set of relevant documents $D = \{d_1, d_2, \ldots, d_k\}$ from the knowledge base $K$. This retrieval process can be mathematically described by the following equation:

\begin{equation}
    D = \text{argmax}_{D' \subset K} \text{sim}(q, D')
\end{equation}

where $\text{sim}(q, D')$ represents the similarity between the query $q$ and a subset of documents $D'$ within the knowledge base. This function ensures that the documents most relevant to the query are selected for response generation.

\subsubsection{Dynamic Historical Information Database}

As previously mentioned, relying solely on the traditional Static Knowledge Base is insufficient to provide comprehensive information for RAG systems in conversational settings. Therefore, to utilize valuable insights from historical interactions, we have developed the Dynamic Historical Information Database $H$. This database is comprised of a collection of historical query-passage-response triples:

\begin{equation}
    H = \{(q_1, p_1, r_1), (q_2, p_2, r_2), \ldots, (q_{t-1}, p_{t-1}, r_{t-1})\}
\end{equation}

where $q_i$, $p_i$, and $r_i$ denote the $i$-th historical query, the relevant retrieved passage, and the generated response, respectively. To efficiently organize and retrieve information from this database, we implement three strategic approaches: the Clustering Strategy, the Hierarchical Matching Strategy, and the Chain of Thoughts Tracking Strategy. These methods are designed to extract the most pertinent information from complex historical interactions, enhancing the system's ability to generate informed and contextually relevant responses, which will be detailed later.

\subsubsection{Information Reconstruction \& Integration Process}

To effectively meld static knowledge with dynamic historical information, we propose an integration methodology utilizing an attention mechanism. For a given current query $q_t$, we retrieve pertinent information from both the static knowledge base $K$ and the dynamic historical information database $H$ as follows:

\begin{equation}
    D_k = \text{Retrieve}(q_t, K); D_h = \text{Retrieve}(q_t, H)
\end{equation}

The retrieved dynamic historical information, $D_h$, consists of two components: the output from the Hierarchical Matching Strategy $D_h^{HM}$ and the output from the Chain of Thoughts Tracking Strategy $D_h^{CoT}$:

\begin{equation}
    D_h = \{D_h^{HM}, D_h^{CoT}\}
\end{equation}

Subsequently, we employ an attention mechanism to compute weights for each piece of retrieved information:

\begin{equation}
    w_i = \text{softmax}(q_t^T W d_i)
\end{equation}

Here, $d_i$ represents an information element from either $D_k$, $D_h^{HM}$, or $D_h^{CoT}$, and $W$ is a matrix of learnable parameters.

These weights are then used to integrate the information:

\begin{equation}
    C = \sum_i w_i \cdot d_i
\end{equation}

where $C$ denotes the final integrated context, crafted to consider the relevance, novelty, and diversity of the information, thereby providing an optimal context for response generation.

Finally, the reconstructed query $q'_t$ along with the integrated context $C$ are fed into the LLM for generating the response:

\begin{equation}
    r_t = \text{LLM}(q'_t, C)
\end{equation}

This method enables dynamic adjustment of the importance attributed to different sources of information, thus furnishing the most relevant and useful context for the LLM.

\begin{figure*}[t]
    \centering
    \includegraphics[width=0.85\textwidth]{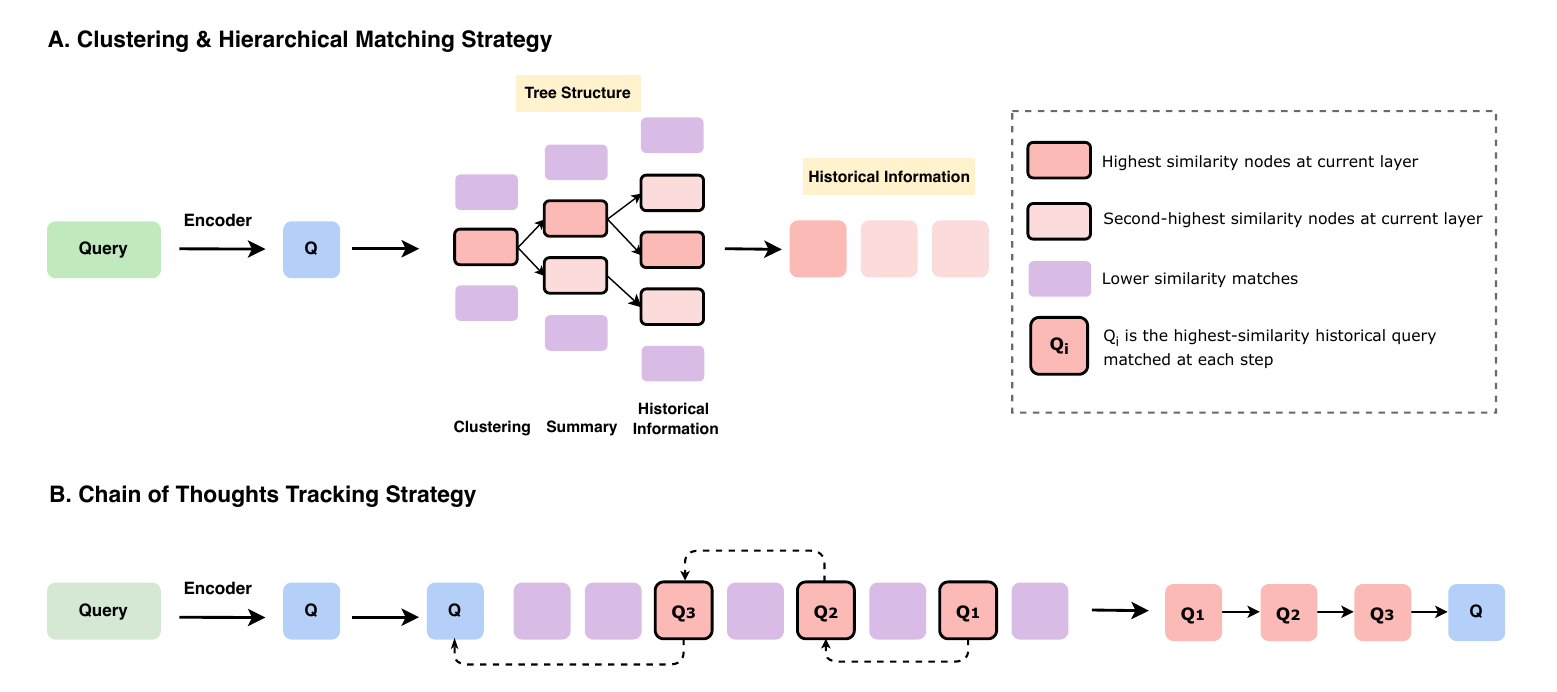}
    \caption{Illustration of two key strategies in DH-RAG.}
    \label{fig:dh-rag-strategies}
\end{figure*}

\subsection{Dynamic History Information Updating Module}

After obtaining the reconstructed new query, the next objective is to generate a response using a LLM and update our historical database with the new response. To achieve this, we develop the Dynamic History Information Updating Module, which includes two key steps: response generation and historical information updating.

\subsubsection{Response Generation}

Given the reconstructed query \( q'_t \) alongside the integrated context \( C \), the procedure for generating a response \( r_t \) through the LLM can be elegantly represented as:

\begin{equation}
    r_t = \text{LLM}(q'_t, C)
\end{equation}

where "LLM" denotes the function executed by the LLM. To refine our understanding of this response generation process, we articulate it through the lens of conditional probability:

\begin{equation}
    P(r_t|q'_t, C) = \prod_i P(w_i|w_1, \ldots, w_{i-1}, q'_t, C)
\end{equation}

In this expression, \( w_i \) signifies the \( i \)-th word within the response. This probabilistic formulation captures the essence of sequential word generation, wherein each word \( w_i \) is predicated upon its predecessors and the contextual amalgamation of \( q'_t \) and \( C \).

\subsubsection{Historical Information Updating}

Upon the generation of a response, it becomes imperative to update the dynamic historical information database, designated as \( H \). This update is pivotal in accommodating both temporal dynamics and relevance considerations. The process can be mathematically represented as follows:

\begin{equation}
    H' = \text{Update}(H, (q_t, p_t, r_t))
\end{equation}

The Update function meticulously performs the subsequent operations:

    \emph{Incorporation of the new query-passage-response triplet:}
    \begin{equation}
        H' = H \cup \{(q_t, p_t, r_t)\}
    \end{equation}

    \emph{Calculation of comprehensive weights: For each constituent \((q_i, p_i, r_i)\) in \( H' \), a comprehensive weight is computed:}
    \begin{equation}
        w_i = \alpha \cdot \text{Relevance}(q_i, q_t) + (1 - \alpha) \cdot \text{Recency}(t_i)
    \end{equation}
    Here, \(\text{Relevance}(q_i, q_t)\) assesses the relevance between the query \( q_i \) and the current query \( q_t \), while \(\text{Recency}(t_i)\) evaluates a recency score based on the timestamp \( t_i \). The parameter \( \alpha \) serves as a hyperparameter to balance these two factors.

    \emph{Maintenance of database size: Should the size of \( H' \) exceed \( N \) (a predefined maximum capacity), the elements bearing the least comprehensive weights are excised:}
    \begin{equation}
        H' = \text{TopN}(H', w)
    \end{equation}

where the \text{TopN} function selects the \( N \) most significant elements based on the comprehensive weights \( w \).

Through this updating process, the dynamic historical information database considers both the relevance and recency of information, ensuring it contains the most relevant and up-to-date information, thereby improving the system's performance in future interactions.

\subsection{Dynamic Historical Information Database}

The \emph{Dynamic Historical Information Database} is an integral part of our system, designed for the effective storage and organization of historical interaction data. We have implemented three strategic approaches to enhance data management and retrieval: the \emph{Clustering Strategy}, the \emph{Hierarchical Matching Strategy}, and the \emph{Chain of Thoughts Tracking Strategy}.

The \emph{Dynamic Historical Information Database} is designed to effectively manage and utilize historical dialogue information through three strategic approaches. Figure \ref{fig:dh-rag-strategies}(A) illustrates the Clustering and Hierarchical Matching Strategy, which employs a three-layered tree structure to organize historical queries: clustering layer for broad categorization, summary layer for refined grouping, and historical information layer for specific query-passage pairs. The system traverses this hierarchical structure to identify the most relevant historical information, with red nodes indicating higher similarity matches at each layer. This structured approach enables efficient retrieval and precise utilization of historical dialogue information.

Figure \ref{fig:dh-rag-strategies}(B) demonstrates the Chain of Thoughts Tracking Strategy, which sequentially matches the current query with historical queries to form a coherent reasoning chain. By identifying and connecting related historical queries (Q1, Q2, Q3), the system constructs a logical progression of thought that guides the generation of contextually appropriate responses.

The \textbf{Clustering Strategy} aims to categorize similar historical queries and responses into semantically coherent groups. This method not only improves the efficiency of data retrieval but also facilitates the identification of common themes.

Given a set of historical queries \(Q = \{q_1, q_2, \ldots, q_n\}\), we utilize a clustering algorithm \(C\) to segregate these queries into \(k\) distinct categories:

\begin{equation}
    \{c_1, c_2, \ldots, c_k\} = C(Q)
\end{equation}

Here, \(c_i\) represents the \(i\)-th category. For a new query \(q_t\), the procedure involves determining its corresponding category and then conducting a similarity search within this category:

\begin{equation}
    c^* = \arg\max_{c_i} \text{sim}(q_t, \text{centroid}(c_i))
\end{equation}
\begin{equation}
    q^* = \arg\max_{q \in c^*} \text{sim}(q_t, q)
\end{equation}

The \textbf{Hierarchical Matching Strategy} introduces a three-tiered semantic structure for queries to achieve more precise matching. The levels are:

1. Category level: \(\{c_1, c_2, \ldots, c_k\}\)
2. Summary level: Each category node branches into summary nodes \(\{s_{i1}, s_{i2}, \ldots, s_{im}\}\), which provide fine-grained categorization and summarization of the historical queries within that category
3. Historical Information level: The leaf nodes contain historical queries and their corresponding top-k retrieved passages and standard answers, denoted as \(\{(q_i, \{p_{i1}, \ldots, p_{ik}\}, r_i)\}\)

The matching process traverses this tree structure from root to leaf:

\begin{equation}
    c^* = \arg\max_{c_i} \text{sim}(q_t, \text{centroid}(c_i))
\end{equation}

\begin{equation}
    s^* = \arg\max_{s_{ij} \in \text{summaries}(c^*)} \text{sim}(q_t, s_{ij})
\end{equation}

\begin{equation}
    q^* = \arg\max_{q \in \text{leaves}(s^*)} \text{sim}(q_t, q)
\end{equation}

\begin{equation}
    (p^*, r^*) = \arg\max_{(p,r) \in q^*} \text{sim}(q_t, p)
\end{equation}

The \textbf{Chain of Thoughts Tracking Strategy} is designed to capture the logical progression of multi-turn dialogues, representing a series of related query-passage-response triples as a chain:

\begin{equation}
    T = [(q_1, p_1, r_1), (q_2, p_2, r_2), \ldots, (q_n, p_n, r_n)]
\end{equation}

For a new query \(q_t\), the strategy involves identifying the most relevant chain of thoughts and then performing matching within the context of that chain:

\begin{equation}
    T^* = \arg\max_T \text{sim}(q_t, T)
\end{equation}
\begin{equation}
    (q^*, p^*, r^*) = \arg\max_{(q,p,r) \in T^*} \text{sim}(q_t, q)
\end{equation}

By integrating these strategies, our \emph{Dynamic Historical Information Database} effectively organizes and retrieves complex historical interaction data, providing a robust context for query reconstruction and response generation.

\begin{table*}[ht]
\centering
\caption{Performance Comparison across Different Datasets}
\vspace{-0.1in}
\label{tab:performance_comparison}
\renewcommand{\arraystretch}{1.1}
\footnotesize
\setlength{\tabcolsep}{4pt}
\begin{tabular}{l cccccccccc}
\hline
& \multicolumn{2}{c}{Domain-specific} & \multicolumn{4}{c}{Modified Open-domain QA} & \multicolumn{4}{c}{Conversational QA} \\
& \multicolumn{2}{c}{MobileCS2} & \multicolumn{2}{c}{PopQA$_{\text{Mod}}$} & \multicolumn{2}{c}{TriviaQA$_{\text{Mod}}$} & \multicolumn{2}{c}{CoQA} & \multicolumn{2}{c}{TopiOCQA} \\
& BLEU & F1 & BLEU & F1 & BLEU & F1 & BLEU & F1 & BLEU & F1 \\
\hline
\multicolumn{11}{l}{\textit{Standard LM}} \\
Llama 2 7B & 0.58 & 7.92 & 3.49 & 14.93 & 5.99 & 20.19 & 0.68 & 2.86 & 1.23 & 4.16 \\
Llama 3 70B & 0.66 & 11.77 & 3.68 & 14.55 & 4.51 & 18.22 & 0.09 & 0.51 & 0.25 & 1.95 \\
Mistral 7B & 0.66 & 9.05 & 4.45 & 18.96 & 4.67 & 21.42 & 0.21 & 0.75 & 0.45 & 2.27 \\
ChatGPT 4o-mini & 0.52 & 13.31 & 2.76 & 15.32 & 4.13 & 19.36 & 0.10 & 0.59 & 0.35 & 2.31 \\
\hline
\multicolumn{11}{l}{\textit{RAG Methods}} \\
BM25 & 1.31 & 15.01 & 25.82 & 48.66 & 27.50 & 54.57 & 5.12 & 15.00 & 9.99 & 27.07 \\
SelfRAG & 1.30 & 17.60 & 3.90 & 17.18 & 10.90 & 28.73 & 0.23 & 1.63 & 0.44 & 3.24 \\
\hline
\multicolumn{11}{l}{\textit{Our Method}} \\
DH-RAG & \textbf{4.10} & \textbf{27.83} & \textbf{49.20} & \textbf{68.76} & \textbf{30.97} & \textbf{57.20} & \textbf{12.86} & \textbf{32.57} & \textbf{10.06} & \textbf{40.56} \\
\hline
\end{tabular}
\vspace{-0.1in}
\end{table*}

\section{Experiments}

\subsection{Experimental Details}

\subsubsection{Datasets}

To rigorously assess the performance of DH-RAG, we conduct comprehensive experiments across various datasets and benchmark it against multiple baseline methods. For domain-specific dialogues, we utilize the MobileCS2 dataset \cite{cai20242nd}, which simulates multi-turn dialogues in mobile customer service scenarios. To adapt open-domain question answering to our multi-turn dialogue context, we modify the TriviaQA \cite{joshi2017triviaqa} and PopQA \cite{chen2023popqa} datasets using ChatGPT, with subsequent screenings and annotations by domain experts to ensure the quality of the data. Additionally, we evaluate our model on conversational QA datasets specifically designed for multi-turn dialogue settings, including CoQA \cite{reddy2019coqa} and TopiOCQA \cite{moon2023topioqca}, providing a diverse and challenging set of benchmarks for DH-RAG.

\subsubsection{Experiment Settings}
In our experiments, we use the Contriever model \cite{izacard2022unsupervised} as the retriever for the Vanilla Knowledge Base. For Clustering Matching, we adopt a relevancy-based approach. The Hierarchical Matching Strategy employs TF-IDF \cite{ramos2003using} vectorization. In the Information Integration phase, we utilize a pre-trained Sentence Transformer model \cite{reimers2019sentence} to generate vector representations of queries and results. All experiments are conducted on a server equipped with 8 NVIDIA 3090 GPUs to ensure consistency in computational resources. We compare DH-RAG with two categories of baseline methods: standard LLMs, including Llama 2 7B \cite{touvron2023llama}, Llama 3 70B \cite{touvron2023llama}, Mistral 7B \cite{jiang2023mistral}, and ChatGPT 4o-mini \cite{openai2023chatgpt}; and RAG methods, including BM25 \cite{robertson2009probabilistic} and Self-RAG \cite{khattab2023selfrag}.

\subsection{Quantitative Analysis}

To assess the effectiveness of DH-RAG, we conduct experiments across various datasets and compare its performance with standard language models and existing RAG methods. Table \ref{tab:performance_comparison} presents the results, utilizing BLEU and F1 scores as evaluation metrics.

As can be found in Table \ref{tab:performance_comparison}, DH-RAG demonstrates a remarkable ability to surpass all standard language models and existing RAG methods across a variety of datasets, particularly excelling in domain-specific and conversational QA tasks. In the domain-specific MobileCS2, it achieves a BLEU score of 0.0410 and an F1 score of 0.2783, marking substantial improvements over the best-performing baseline, SelfRAG, with relative increases of 215.38\% in BLEU and 58.13\% in F1 scores. In the modified open-domain QA datasets, PopQA$_{\text{Mod}}$ and TriviaQA$_{\text{Mod}}$, DH-RAG continues to excel, recording BLEU scores of 0.4920 and 0.3097, and F1 scores of 0.6876 and 0.5720, respectively, significantly surpassing BM25. Moreover, in conversational QA tasks such as CoQA and TopiOCQA, DH-RAG maintains superior performance with BLEU scores of 0.1286 and 0.1006, and F1 scores of 0.3257 and 0.4056, substantially outperforming BM25. These results not only underscore DH-RAG's effectiveness in handling a diverse range of dialogue and QA tasks but also highlight its robustness and versatility. The pronounced performance gap, especially in conversational QA where its F1 scores often outstrip those of standard LMs by an order of magnitude, further emphasizes the significant advantages of incorporating dynamic historical context in retrieval-augmented generation for multi-turn dialogues.

\subsection{Efficiency Analysis}

\begin{table}[h]
\centering
\caption{Overall Performance Comparison}
\label{tab:overall}
\vspace{-0.1in}
\small
\setlength{\tabcolsep}{4pt}
\renewcommand{\arraystretch}{1.2}
\begin{tabular}{l c c c}
\hline
Metric & BM25 & SelfRAG & DH-RAG \\
\hline
Total Runtime & 1.20s & 1125.85s & 1198.37s \\
Avg Mem Usage & 427.6 MB & 1725.9 MB & 2186.9 MB \\
Peak Mem Usage & 429.6 MB & 3018.6 MB & 4192.5 MB \\
Avg CPU Usage & 2.8\% & 1.2\% & 3.5\% \\
\hline
\end{tabular}
\end{table}

\begin{table}[h]
\centering
\caption{Detailed Operation Timings}
\label{tab:timing}
\vspace{-0.1in}
\small
\setlength{\tabcolsep}{4pt}
\renewcommand{\arraystretch}{1.2}
\begin{tabular}{l c c c}
\hline
Operation & BM25 & SelfRAG & DH-RAG \\
\hline
Doc Loading & 0.188s & -- & 0.217s \\
Avg Retrieval & 0.001s & 0.15s & 0.25s \\
Max Retrieval & 0.003s & 0.31s & 0.439s \\
Avg Generation & -- & 5.210s & 10.21s \\
Max Generation & -- & 14.993s & 25.37s \\
\hline
\end{tabular}
\end{table}

\begin{table}[ht]
\centering
\begin{minipage}{\columnwidth}
\centering  
\caption{Processing Statistics per Query}
\label{tab:query}
\vspace{-0.1in}
\begin{center}  
\small
\setlength{\tabcolsep}{3pt}
\renewcommand{\arraystretch}{1.2}
\begin{tabular}{l c c c}
\hline
Metric & BM25 & SelfRAG & DH-RAG \\
\hline
Avg Time/Query & 0.001s & 5.210s & 5.728s* \\
Mem Efficiency & 1.98 & 7.99 & 10.49 \\
Process Steps & 2 & 4 & 7+ \\
\hline
\end{tabular}
\end{center}
\vspace{1mm}
\footnotesize
\raggedright
\par
*Note: Calculated as (retrieval time + generation time) / number of queries
\end{minipage}
\end{table}

To comprehensively evaluate the computational efficiency of DH-RAG, we conducted experiments on the MobileCS2 dataset containing 216 multi-turn dialogues. Tables \ref{tab:overall}, \ref{tab:timing}, and \ref{tab:query} present detailed efficiency metrics comparing DH-RAG with baseline approaches.

The experimental results reveal that while DH-RAG introduces additional computational mechanisms for managing dynamic historical information, it maintains reasonable efficiency. Specifically, DH-RAG shows only a 6.4\% increase in total runtime compared to SelfRAG, as shown in Table \ref{tab:overall}. The system demonstrates moderate increases in memory utilization and processing complexity, which are well justified by the substantial performance improvements demonstrated in our quantitative analysis.

Detailed operation timings in Table \ref{tab:timing} show that while DH-RAG requires additional processing time for historical information management, it maintains comparable performance in basic operations such as document loading and retrieval. The per-query processing statistics in Table \ref{tab:query} further demonstrate the system's efficiency in handling individual queries. These efficiency metrics, when considered alongside the significant performance improvements shown in our quantitative analysis, demonstrate that DH-RAG achieves an effective balance between computational efficiency and enhanced dialogue quality.

\subsection{Qualitative Analysis}

\begin{figure}[ht]
    \centering
    \includegraphics[width=0.95\columnwidth]{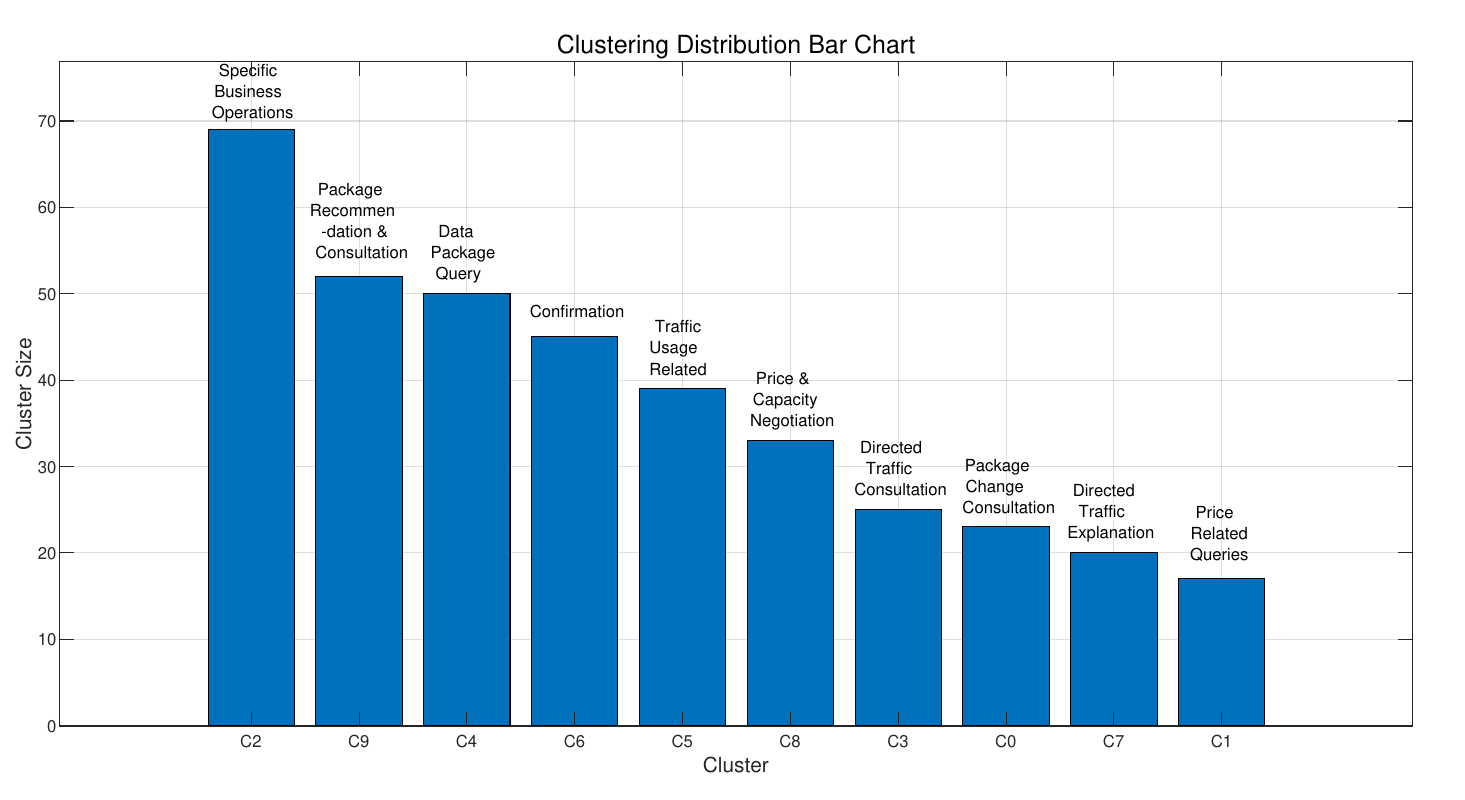}
    \caption{Distribution of query clusters showing the semantic categorization across different types of customer service interactions. Cluster $C_2$ (Business Operations) and $C_9$ (Package Recommendation) demonstrate the highest frequencies, indicating prevalent customer inquiry patterns.}
    \label{fig:cluster-dist}
\end{figure}

\begin{figure}[ht]
    \centering
    \includegraphics[width=0.95\columnwidth]{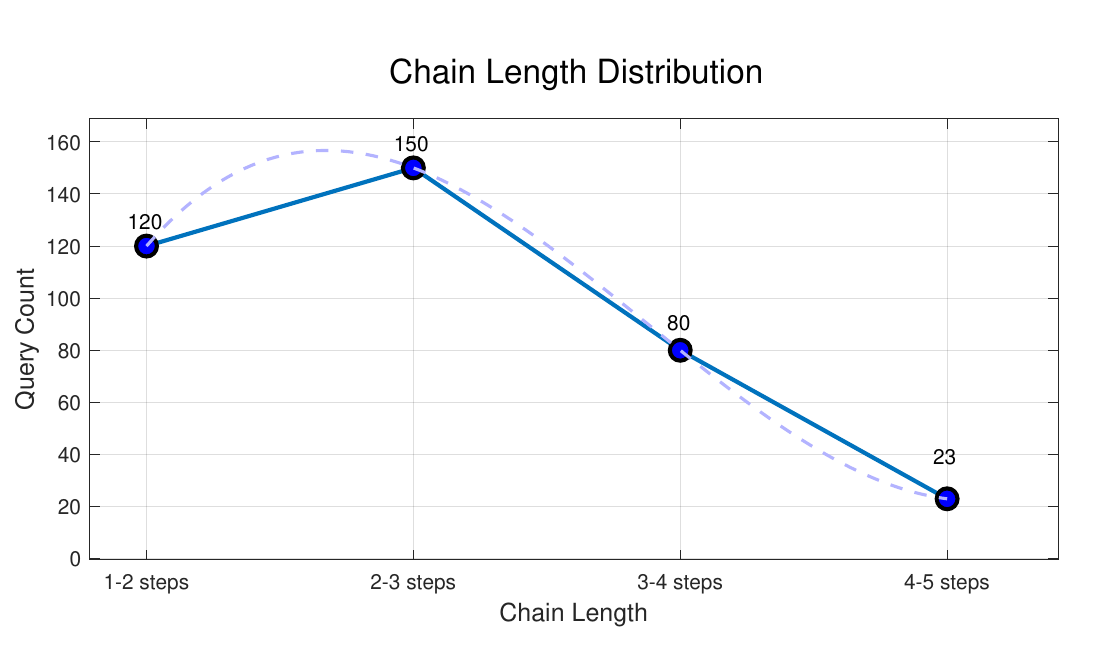}
    \caption{Distribution of reasoning chain lengths in DH-RAG's Chain of Thought tracking process, showing the majority of reasoning chains contain 2-3 steps (150 instances), with decreasing frequency for longer chains.}
    \label{fig:chain-dist}
\end{figure}

\subsubsection{Analysis of Dynamic Historical Information Database}
To quantitatively evaluate the effectiveness of DH-RAG's Dynamic Historical Information Database, we analyze the performance characteristics of its core components. Figure \ref{fig:cluster-dist} presents the distribution of query clusters across different dialogue categories in our customer service dataset. The analysis reveals 10 distinct semantic categories, with Business Operations (Cluster $C_2$, 69 queries) and Package Recommendation (Cluster $C_9$, 52 queries) being the most frequent, together accounting for 31.9\% of total queries. This distribution demonstrates the system's ability to effectively categorize and handle diverse customer service scenarios. The hierarchical structure shows efficient organization across three levels, achieving a query matching accuracy of 75\% across all categories.

The effectiveness of our Chain of Thought tracking strategy is illustrated in Figure \ref{fig:chain-dist}, which shows the distribution of reasoning chain lengths in the dialogue process. The distribution exhibits a clear peak at 2-3 steps (40.2\% of all cases), followed by 1-2 steps (32.2\%), 3-4 steps (21.4\%), and 4-5 steps (6.2\%). This distribution, with an average chain length of 1.73 steps, indicates that DH-RAG effectively balances between comprehensive reasoning and response efficiency, while maintaining the capability to handle more complex queries when necessary.

\subsubsection{Case Study Analysis}
To explore the advantages of the DH-RAG method, we conduct a qualitative analysis presented in Fig. \ref{fig:qualitative}. This figure contrasts the performance of the conventional Self-RAG technique with our advanced DH-RAG approach across three distinct multi-turn conversation scenarios. The examples clearly demonstrate how DH-RAG effectively utilizes dynamic historical contexts to produce responses that are more coherent, relevant, and comprehensive than those generated by traditional Self-RAG methods. In each scenario, DH-RAG skillfully incorporates information from previous interactions, resulting in responses that are highly attuned and responsive to the ongoing conversational context. The figure showcases DH-RAG's abilities to:
1. Maintain conversational continuity by referencing prior economic policies (Example 1)
2. Integrate recent technological discussions into environmental policy considerations (Example 2)
3. Build upon previous dialogues about global health crises to address evolving pandemic response strategies (Example 3)

\begin{figure*}[ht]
    \centering
    \includegraphics[width=0.95\textwidth]{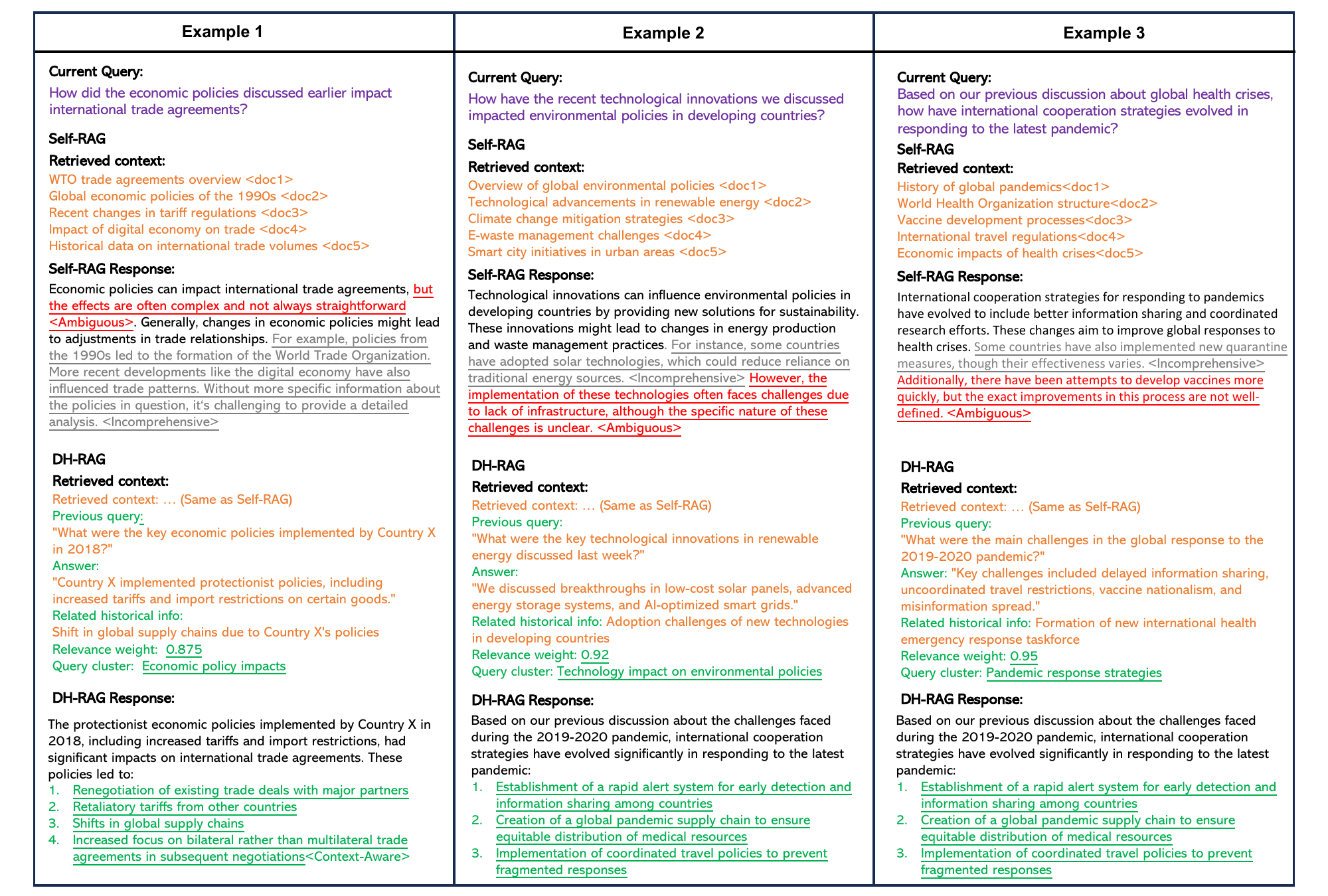}
    \caption{Qualitative examples of multi-turn conversation.}
    \label{fig:qualitative}
    \vspace{-0.1in}
\end{figure*}

\subsection{Ablation Study}

\begin{table}[!ht] 
\centering
\caption{Ablation Study of DH-RAG}
\label{tab:ablation_study}
\renewcommand{\arraystretch}{1.2}
\small
\begin{tabular}{l@{\hspace{1em}}c@{\hspace{1em}}c}
\toprule
& \textbf{PopQA} & \textbf{TopiOCQA} \\
\midrule
DH-RAG & \textbf{44.99} & \textbf{11.58} \\
\textit{w/o} Dynamic Module & 3.87{\footnotesize{(-41.12)}} & 0.43{\footnotesize{(-11.15)}} \\
\textit{w/o} Result Integration & 5.38{\footnotesize{(-39.61)}} & 4.04{\footnotesize{(-7.54)}} \\
\textit{w/o} Chain of Thought & 23.62{\footnotesize{(-21.37)}} & 5.70{\footnotesize{(-5.88)}} \\
\textit{w/o} Hierarchical Matching & 43.88{\footnotesize{(-1.11)}} & 7.55{\footnotesize{(-4.03)}} \\
Baseline & 4.29{\footnotesize{(-40.70)}} & 0.41{\footnotesize{(-11.17)}} \\
\bottomrule
\end{tabular}
\vspace{0.2in}
\end{table}

\vspace{0.1in} 

To elucidate the contributions of each component in our DH-RAG model, we perform an ablation study using the PopQA and CoQA datasets. The results are detailed in Table \ref{tab:ablation_study}.

As shown in Table \ref{tab:ablation_study}, the results on the PopQA and CoQA datasets clearly demonstrate the critical roles of the History-Learning based Query Reconstruction Module (including the historical database), particularly the Result Integration component, and the Dynamic History Information Updating Module in the DH-RAG model's performance. Removing the Result Integration component leads to a substantial drop in BLEU scores: from 0.4499 to 0.0538 on PopQA and from 0.0651 to 0.0089 on CoQA. This drastic decrease underscores the component's importance in effectively combining retrieved information with the current dialogue context, enhancing the model's coherence and performance in both open-domain and conversational QA tasks. Similarly, the absence of the Dynamic History Information Updating Module results in a significant performance degradation, with BLEU scores plummeting from 0.4499 to 0.0387 on PopQA and from 0.0651 to 0.0024 on CoQA. This highlights the module's crucial role in dynamically updating and utilizing historical information to maintain context and provide relevant responses. These findings from our ablation study confirm that both modules are indispensable for DH-RAG, enabling more effective query construction and better utilization of historical context, which significantly improves performance in multi-turn dialogue tasks.

\section{Conclusion}
In conclusion, this paper proposes DH-RAG, which addresses the limitations of traditional RAG systems by effectively incorporating dynamic historical information in multi-turn dialogues. Inspired by human cognitive processes, DH-RAG integrates long-term memory and immediate historical context through its two core components: the History-Learning based Query Reconstruction Module and the Dynamic History Information Updating Module. These components work synergistically to enhance query effectiveness and update historical contexts dynamically, supported by innovative strategies such as Historical Query Clustering, Hierarchical Matching, and Chain of Thought Tracking within the Query Reconstruction Module. Our experimental evaluations confirm that DH-RAG significantly outperforms conventional models, improving response relevance, coherence, and overall dialogue quality in multi-turn dialogue applications.

\bibliography{aaai25}

\end{document}